\documentclass[letterpaper]{article} % DO NOT CHANGE THIS
\usepackage{aaai24}  % DO NOT CHANGE THIS
\usepackage{times}  % DO NOT CHANGE THIS
\usepackage{helvet}  % DO NOT CHANGE THIS
\usepackage{courier}  % DO NOT CHANGE THIS
\usepackage[hyphens]{url}  % DO NOT CHANGE THIS
\usepackage{graphicx} % DO NOT CHANGE THIS
\usepackage{natbib}  % DO NOT CHANGE THIS AND DO NOT ADD ANY OPTIONS TO IT
\usepackage{caption} % DO NOT CHANGE THIS AND DO NOT ADD ANY OPTIONS TO IT
\usepackage[utf8]{inputenc}
\usepackage{lipsum}
\usepackage{amssymb,amsthm,amsmath}
\usepackage{enumitem}
\usepackage{caption}
\usepackage{subcaption}
\usepackage{centernot}
\usepackage[ruled,vlined,linesnumbered]{algorithm2e} % noend or vlined to remove end
\usepackage{mathrsfs}   % \mathscr fonts
\usepackage{mathtools}  % for prescript
\usepackage{tabularx}   % adjustable table width
\usepackage{multirow}   % multiple rows and columns
\usepackage{array}      % allows m{'width'} and b{'width'}
\usepackage{booktabs}   % \toprule \midrule \bottomrule
\usepackage{colortbl}   % allows table cells to be coloured
\usepackage{xargs}      % Use more than one optional parameter in a new commands
\usepackage{tikz}
\usepackage{tikz-cd}
\usepackage{todonotes}
\usepackage{soul}       % underline with line wrapping (\ul)
\usepackage{marginnote} % https://tex.stackexchange.com/questions/297684

\newcommand{\colourDZC}{ao(english)}  
  
\newcommand{\todocustom}[3]{\todo[linecolor=#2,backgroundcolor=#2!25,bordercolor=#2,#3]{#1}}  
\newcommandx{\tododzc}[1]{{\color{\colourDZC}{#1}}}  
\newcommandx{\dzc}[1]{{\color{\colourDZC}{#1}}}  
\newcommandx{\todoDZC}[2][1=]{\todocustom{#2}{\colourDZC}{#1}}

\definecolor{caribbeangreen}{rgb}{0.0, 0.8, 0.6}
\definecolor{brilliantlavender}{rgb}{0.96, 0.73, 1.0}
\definecolor{amethyst}{rgb}{0.6, 0.4, 0.8}
\definecolor{ao(english)}{rgb}{0.0, 0.5, 0.0}
\definecolor{arylideyellow}{rgb}{0.91, 0.84, 0.42}
\definecolor{asparagus}{rgb}{0.53, 0.66, 0.42}
\definecolor{aquamarine}{rgb}{0.5, 1.0, 0.83}
\definecolor{babyblue}{rgb}{0.54, 0.81, 0.94}
\definecolor{fwtchanged}{rgb}{0.3, 0.3, 0.7}
\definecolor{rosewood}{rgb}{0.4, 0.0, 0.04}
\definecolor{oldmauve}{rgb}{0.4, 0.19, 0.28}
\definecolor{myrtle}{rgb}{0.13, 0.26, 0.12}
\definecolor{magenta(dye)}{rgb}{0.79, 0.08, 0.48}

\newcommand{\sylvie}[1]{{\color{amethyst} #1}}

\DeclareMathOperator*{\pre}{pre}

\DeclareMathOperator*{\eff}{eff}

\newtheorem{theorem}{Theorem}[section]

\theoremstyle{definition}
\newtheorem{definition}[theorem]{Definition}

\def\N{\mathbb{N}}
\def\R{\mathbb{R}}

\def\g{\gamma}

\renewcommand{\phi}{\varphi}

\newcommand{\abs}[1]{\left| #1 \right|}

\newcommand{\gen}[1]{\left< #1 \right>}

\newcommand{\set}[1]{\left\{ #1 \right\}}

\newcommand{\brk}[1]{\left[ #1 \right]}

\newcommand{\Biglr}[1]{\Bigl( #1 \Bigr)}

\DeclareMathOperator*{\concat}{%
  {\Vert}%
}

\newcommand{\header}[1]{\multicolumn{1}{r}{\rotatebox[origin=l]{90}{\hspace*{-0.22cm} #1}}}
\newcommand{\colorofcell}{blue}
\newcommand{\cellfontsize}{\scriptsize}
\newcommand{\first}[1]{\cellfontsize\cellcolor{\colorofcell!30}{\textbf{#1}}}
\newcommand{\second}[1]{\cellfontsize\cellcolor{\colorofcell!20}{{#1}}}
\newcommand{\third}[1]{\cellfontsize\cellcolor{\colorofcell!10}{{#1}}}
\newcommand{\normalcell}[1]{\cellfontsize{{#1}}}
\newcommand{\zerocell}[1]{\cellfontsize{-}}

\newcommand{\width}{\prescript{k}{}{\text{PN}}_\novFeat^h}
\newcommand{\qb}{\prescript{k}{}{\text{QB}}_\novFeat^h}
\newcommand{\qbSumDefn}{\displaystyle\sum}
\newcommand{\novFeat}{f}
\newcommand{\novHeur}{n}
\newcommand{\seqSet}[1]{{#1^{<\N}}}
\newcommand{\seqState}[1]{\gen{#1}}

\newcolumntype{Y}{>{\raggedleft\arraybackslash}X}
\newcolumntype{Z}{>{\centering\arraybackslash}X}

\newcommand{\hgc}{h^{\text{gc}}}
\newcommand{\gc}{\hgc}
\newcommand{\hgdc}{h^{\text{md}}}
\newcommand{\aibr}{h^{\text{aibr}}}
\newcommand{\hadd}{h^{\text{add}}}
\newcommand{\hradd}{h^{\text{radd}}}
\newcommand{\hmrp}{h^{\text{mrp}}}
\newcommand{\hj}{\text{\scriptsize{+hj}}}
\newcommand{\hmrphj}{\hmrp\hj}

\newcommand{\A}{\text{A}}
\newcommand{\B}{\text{B}}
\newcommand{\I}{\text{I}}
\newcommand{\PN}{\text{PN}}
\newcommand{\QB}{\text{QB}}
\newcommand{\SQ}{\text{S}}
\newcommand{\MQ}{\text{M}}
\newcommand{\PF}{\text{P}}
\newcommand{\MQu}{\MQ}
\newcommand{\PFo}{\PF}

\newcommand{\hwadd}{\hadd_{\gen{\A,\PN}}}
\newcommand{\hiwadd}{\hadd_{\gen{\B,\PN}}}
\newcommand{\hqbadd}{\hadd_{\gen{\A,\QB}}}
\newcommand{\hiqbadd}{\hadd_{\gen{\B,\QB}}}
\newcommand{\hiqbgdc}{\hgdc_{\gen{\B,\QB}}}
\newcommand{\hiqbmrphj}{\hmrp_{\gen{\B,\QB}}\hj}

\newcommand{\patty}{\textsc{Patty}}

\newcommand{\mqe}{\MQu({3h})}
\newcommand{\mqeiqb}{\MQu(3n)}
\newcommand{\mqeeiqb}{\MQu(3h\!\concat\!3n)}

\newcommand{\pfh}{\PFo(3h)}
\newcommand{\pfn}{\PFo(3n)}
\newcommand{\pfhn}{\PFo(3h\!\concat\!3n)}

\newif\iflargermidtables
\newif\ifrecursive 
\title{Novelty Heuristics, Multi-Queue Search, and Portfolios for Numeric Planning}
\author{
    Dillon Z. Chen$^{1,2}$, Sylvie Thi\'ebaux$^{1,2}$
}
\affiliations{
    $^{1}$LAAS-CNRS, Universit\'e de Toulouse\\
    $^{2}$Australian National University\\
    \{dillon.chen, sylvie.thiebaux\}@laas.fr
}

\newcommand{\hgdcname}{Manhattan distance}

\begin{document}
\maketitle

\begin{abstract}
Heuristic search is a powerful approach for solving planning problems and numeric planning is no exception. In this paper, we boost the performance of heuristic search for numeric planning with various powerful techniques orthogonal to improving heuristic informedness: numeric novelty heuristics, the \hgdcname{} heuristic, and exploring the use of multi-queue search and portfolios for combining heuristics.
\end{abstract}

\section{Introduction}
Numeric planning is an expressive extension of classical planning where states are able to exhibit real-valued variables. 
It was formalised in PDDL2.1~\cite{fox:long:jair03} and is undecidable in the general case~\cite{helmert:aips02}. 
However, it is PSPACE-complete when variables are integer and bounded, and there also exist compilations from numeric to classical planning with certain features that preserve plan length~\cite{gigante:scala:ijcai23}.
Similarly to classical planning, the two main approaches for solving numeric planning problems in the literature consist in either heuristic search or compilation into a sequence of satisfiability problems for constraint-based solvers.
A recent SAT Modulo Theory (SMT) approach, \patty{}, has been proposed~\cite{cardellini:etal:aaai24} which provably encodes fewer variables and constraints than previous constraint-based approaches~\cite{scala:etal:icaps16,bofill:etal:ijcai17}.
Compared to heuristic search, it provides better performance on highly numeric planning problems which do not require long plans.

On the other hand, the state of the art for heuristic search in numeric planning still consists of a search with a single heuristic derived from a relaxation of the problem.
Such heuristics include the numeric extensions of classical planning heuristics: metric FF~\cite{hoffmann:ecai02,hoffmann:jair03}, LP-based heuristics for computing tighter bounds on the relaxed planning graph~\cite{coles:etal:icaps08}, interval-based, subgoaling, and set-additive derivations of $\hadd$~\cite{scala:etal:ijcai16,scala:etal:ecai16,scala:etal:jair20,scala:etal:icaps20}, landmarks~\cite{scala:etal:ijcai17}, admissible IP-based heuristics~\cite{piacentini:etal:aaai18}, LM-cut and operator counting heuristics~\cite{kuroiwa:etal:icaps21,kuroiwa:etal:jair22,kuroiwa:etal:ecai23}.
Alternatively, \citet{illanes:mcilraith:ijcai17} use a partial policy computed from an abstraction of numeric planning into classical planning for search guidance.
Other techniques orthogonal to defining new heuristics for numeric planning also exist, such as abstracting linear numeric problems into simple numeric problems,
enabling the use of simple numeric heuristics~\cite{li:etal:ijcai18}, and computing state space symmetries with the numeric problem description graph~\cite{shleyfman:etal:icaps23}.

In this paper, we boost the performance of heuristic search for numeric planners in several orthogonal directions by porting search techniques from classical planning.
This includes (1) formalising novelty heuristics for numeric planning, (2) constructing the simple but effective \hgdcname{} heuristic, (3) extending the ENHSP planner~\cite{scala:etal:icaps20} with multi-queue search, and (4) building naive portfolios.
We provide a systematic set of experiments that demonstrate the effectiveness of these approaches on the Numeric Track of the 2023 International Planning Competition (NT-IPC)~\cite{arxer:scala:23} and highlight the complementary nature of heuristic search and constraint-based approaches for numeric planning.

\section{Background}
A numeric planning task~\cite{fox:long:jair03} is given by a tuple $\Pi = \gen{X_p, X_n, A, s_0, G}$ where $X_p$ is a finite set of Boolean variables with domains $\set{\top, \bot}$ and $X_n$ is a finite set of numeric variables with domains $\R$. 
Let $X = X_p \cup X_n$ denote the set of state variables. 
Let $S$ denote the set of possibly infinitely many states in $\Pi$, where a state is a total assignment to Boolean and numeric variables. 
Let $s_0\in S$ be the initial state.
A propositional condition is a positive literal, and a numeric condition has the form $\xi \unrhd 0$ where $\xi$ is an arithmetic expression over numeric variables and $\unrhd \in \set{\geq, >, =}$. 
Let $[x]^s$ and $[\xi]^s$ denote the value of $x$ and $\xi$ in $s$, respectively. 
The goal condition $G$ is a set of propositional and numeric conditions, and we say that a state $s$ satisfies $G$ if $s$ satisfies all conditions in $G$.
An action $a$ consists of preconditions and effects.
Action preconditions $\pre(a)$ are sets of propositional and numeric conditions, and action effects $\eff(a)$ assign Boolean values to Boolean variables, and assign the value of a numeric variable using an arithmetic expression.
An action $a$ is applicable in a state $s$ if $s$ satisfies $\pre(a)$, in which case its successor $a(s)$ is the state where the effects $\eff(a)$ are applied to the variables in state $s$.
If $a$ is not applicable in $s$, we set $a(s) = s_\bot\not\in S$.
A plan for a numeric planning task is a sequence of actions $a_1, \ldots, a_n$ such that $s_i = a_i(s_{i-1}) \not= s_{\bot}$ for all $1 \leq i \leq n$ and $s_n$ satisfies $G$.
A numeric planning task is solvable if there exists a plan for it.

\section{Boosting Search for Numeric Planning}
In this section, we outline various techniques to boost the performance of satisficing numeric planning solvers.
Firstly, we extend and combine novelty heuristics into a single framework for numeric planning.
We then introduce a simple extension of the goal count heuristic for numeric planning which outperforms more sophisticated numeric planning heuristics for single queue search.
Lastly, we make use of multi-queue search and portfolios for numeric planning.

\subsection{Numeric Novelty Heuristics}
Let $\seqSet{S}$ denote the set of all finite sequences of states in a numeric planning task, and $\seqState{T}$ a finite sequence of states. 
%
%The notation
\ifrecursive
Let
\else
We write $T_k$ for the state of index $k$ in $\seqState{T}$, and $\seqState{T}.s$ for the sequence obtained by appending state $s$ at the end of $\seqState{T}$. Moreover, let
\fi
$\seqState{T}[:s]$ stand for the substring of $\seqState{T}$ that ends in the element before $s$, and $\seqState{T_{=h(s)}}$ for the substring of $\seqState{T}$ consisting of the states $s'$ such that $h(s')=h(s)$, where $h$ is a heuristic function.
We also assume an enumeration $x_1, \ldots, x_{N}$ of variables in $X$, where $N=\abs{X}$.

\begin{definition}[Novelty Feature]
    A \emph{novelty feature} is a function $\novFeat: \seqSet{S} \times S \to (\R \cup \set{\bot})^{N}$.
\end{definition}

We provide two examples of novelty features for numeric planning. 
The first example is the \emph{assignment feature} $\A$, which has been explored in the context of control systems~\cite{ramirez:etal:aamas18}. 
The function maps a state to a vector of its assignments.
More specifically, the $i$-th element of $\A(\seqState{T}, s)$ is the value of $x_i$ in $s$ if $x_i$ is a numeric variable and otherwise if $x_i=\top$ the element is 1 and else $\bot$.
Note that the function is agnostic to the $\seqState{T}$ input.

\ifrecursive
\else

The next example is the \emph{boundary extension encoding feature}~\cite{teichteil:etal:ijcai20} $\B$, which maps a state to a vector whose component for a particular variable reflects the number of times this variable reached a new extreme value in the past before reaching its current value. 
More formally, for a variable $x_i$ and state $s$ such that $[x_i]^s > [x_i]^{s_0}$, the novelty feature $I(\seqState{T},s,i)$ is the number of times $x_i$ is strictly greater than its value in all preceding states in $T$ before reaching a state $s_j$ where $[x_i]^{s_j} \geq [x_i]^{s}$.
That is
\begin{align}
  I(\seqState{T},s,i) = \sum_{k=1}^{J_I} \Biglr{\bigwedge_{l=1}^{k-1} [x_i]^{T_k} > [x_i]^{T_l}},
\end{align}
where $J_I \leq |\seqState{T}|$ is the smallest index $j$ such that $[x_i]^{T_j} \geq [x_i]^{s}$, and the boolean expression evaluates to 1 or 0.
The case where $[x_i]^s < [x_i]^{s_0}$ is similar, but we now look at decreasing values of $x_i$ with
\begin{align}
  D(\seqState{T},s,i) = \sum_{k=1}^{J_D} \Biglr{\bigwedge_{l=1}^{k-1} [x_i]^{T_k} < [x_i]^{T_l}},
\end{align}
where $J_D \leq |\seqState{T}|$ is the smallest index $j$ such that $[x_i]^{T_j} \leq [x_i]^{s}$.
We can then define the boundary extension encoding feature $B$ on its individual component variables
\begin{align}
    \B(\seqState{T}, s)_i = 
    \begin{cases}
      0, &\text{if $[x_i]^s = [x_i]^{s_0}$} \\
      I(\seqState{T}.s,s,i), &\text{if $[x_i]^s > [x_i]^{s_0}$} \\
    - D(\seqState{T}.s,s,i), &\text{if $[x_i]^s < [x_i]^{s_0}$.}     
    \end{cases}
\end{align}

\fi

Next, we move on to defining a novelty heuristic given a novelty feature and a base heuristic.
We note that the definition can be generalised to take as input a tuple of heuristics but we omit this for simplicity of notation.
\begin{definition}[Novelty Heuristic]
    Given a novelty feature $f$ and a heuristic $h$, a \emph{novelty heuristic} is a function $\novHeur_{\novFeat}^h: \seqSet{S} \times S \to \R$.
\end{definition}

We again provide two examples of novelty heuristics.
The first is the \emph{partition novelty} ($\PN$) measure heuristic.
It was first introduced for solving planning problems in polynomial time with an incomplete search~\cite{lipovetzky:geffner:ecai12} and later adapted for heuristic search~\cite{lipovetzky:geffner:aaai17,correa:seipp:icaps22}.
Given an integer $k \geq 1$, we extend its definition to numeric planning by the function $\width$ where $\width(\seqState{T}, s)$ is the smallest $n \leq k$ for which there is an $n$-subset of variable indices $i_1, \ldots, i_n$ such that none of $x_{i_j}=\bot$ in $s$ and $f(\seqState{T_{=h(s)}}, s)[i_1,\ldots,i_n]$ differs from $f(\seqState{T_{=h(s)}}\brk{:s'}, s')[i_1,\ldots,i_n]$ for all $s' \in \seqState{T_{=h(s)}}$, and $k+1$ if no such subset exists.
The original partition novelty operates on a tuple of heuristics and it is easy to extend this definition to do so.

Another example of a novelty heuristic is the \emph{quantified both} $(\QB)$ novelty function~\cite{katz:etal:icaps17}.
The weakness of the partition novelty measure is that its value does not prioritise states with the same partition novelty based on the base heuristic value.
For example, two states with heuristic values 5 and 3 and the same partition novelty measure will be treated equally in the search, unless a tie-breaking strategy is used.
Furthermore, non-novel states are also treated the same.
The quantified both novelty function addresses these issues by making use of the base heuristic value and also distinguishing non-novel states.
We extend the quantified both novelty function to the numeric case and also to operate on $k$-subsets of variables instead of just single facts.

With abuse of notation, we omit $\gen{T}$ and the novelty feature $f$ from various helper function arguments. 
Firstly, we quantify the novelty of subsets of variables by the function
\begin{align}
    N(J, s)=\min_{s' \in T, f(\gen{T},s)_i=f(\gen{T},s')_i, \forall i \in J} h(s') \label{eq:qbN}
\end{align}
where $\min$ returns $\infty$ if the set it operates over is empty.
The function returns the minimum value of the heuristic over the past states that share the same features as $s$ for the considered variable subset.
\newcommand{\subsets}{\genfrac{\lbrace}{\rbrace}{0pt}{}}
Then we define $\phi(J,s) = h(s)<N(J,s)$ and $\psi(J,s)=h(s)>N(J,s)$ where the Boolean output evaluates to 0 or 1.
The function $\phi$ states that the $J$ values of $s$ are novel if $h(s)$ is strictly lower than the heuristic value of all previously seen states with the same $J$ values.
Similarly, $\psi$ denotes non-novelty by noting whether the heuristic value is strictly greater than the minimum heuristic value of previously seen states with the same $J$ values.
Let $\subsets{N}{n}$ be the set of $n$-subsets of variable indices of $s$ and ${N}\choose{n}$ its cardinality.
Given an integer $k \geq 1$, we can define the quantified both novelty function $\qb$ by 
{
\begin{align}
    &\hspace*{1cm}\qb(\seqState{T}, s) = \notag\\
    &\begin{cases}
        {\qbSumDefn_{n=1}^1} {{N}\choose{n}} - {\qbSumDefn_{J \in \subsets{N}{1}}} \phi(J, s), & \text{if }\exists J \in \subsets{N}{1}, \phi(J, s) \\[-0.3em]
        \hspace*{1.85cm}\vdots & \\[-0.7em]
        {\qbSumDefn_{n=1}^k} {{N}\choose{n}} - {\qbSumDefn_{J \in \subsets{N}{k}}} \phi(J, s), & \text{if }\exists J \in \subsets{N}{k}, \phi(J, s) \\[-0.5em]
        {\qbSumDefn_{n=1}^k} {{N}\choose{n}} + {\qbSumDefn_{J \in \subsets{N}{k}}} \psi(J, s), & \text{otherwise,}
    \end{cases}
\end{align}
}%
where cases are tie-broken by priority from the top.
Unlike the partition novelty measure, it is not as straightforward to define the quantified both novelty function for tuples of heuristics.
This is because Eqn.~\eqref{eq:qbN} would return a Pareto set for the $\min$ function if multiple heuristic were used.
One would then require defining $\phi(J, s)$ over Pareto sets in which case there are several ways to do so.

\subsection{Manhattan Distance Heuristic}
The \emph{goal count heuristic} $\hgc$ for classical planning is the simplest non-degenerate heuristic which counts the number of achieved goal propositions in a state. 
It has been extended to numeric planning by counting the number of achieved propositional and numeric goal conditions in the current state.
However, we can easily refine the heuristic for numeric conditions $\xi \unrhd 0$ by measuring the error of the expression $\xi$ evaluated in the current state. 
To formally define this more refined goal count heuristic, let $G=G_p \cup G_n$ be the goal of our problem where $G_p$ is the set of propositional conditions and $G_n$ the set of numeric conditions. 
Let $[l]^s=1$ if $s$ satisfies the literal $l$ and $0$ otherwise. 
Then, given a goal condition $G$ we define the \emph{Manhattan distance heuristic} $\hgdc$ by 
\begin{align}
    \hgdc(s) = \sum_{l \in G_p} [l]^s + \sum_{c=\xi \unrhd 0 \in G_n} \g(c, s)
\end{align}
where the error is $\g(c, s) = 0$ if $s$ satisfies $c$ and $\abs{[\xi]^s}$ otherwise.
It is possible to further refine $\hgdc$ by dividing the error of each numeric condition $\xi \unrhd 0$ by some constant computed from the set of actions relevant to $\xi$. 
For example, we could choose the constant to be either the min or max action effect that brings $\abs{\xi}$ closer to 0 if the condition is not yet satisfied.

\subsection{Multi-Queue Search}
Multi-queue search is an effective method for combining heuristics for satisficing search by maintaining a separate queue in greedy best first search for each heuristic~\cite{helmert:06:jair,roger:helmert:icaps10}.
More specifically, given $n$ heuristics, we have $n$ priority queues from which we pop nodes to expand in a round-robin manner.
After a node is expanded, the children are evaluated by each heuristic and inserted into the corresponding priority queues.
In our implementation, we assume no node reopening which means states are expanded at most once over all queues, at the cost of potentially lower quality solutions.
Naturally, when a novelty heuristic and its base heuristic are used in the same multi-queue search,
we can save computing the base heuristic's value twice.
We note that multi-queue search is not limited to using heuristic values for determining queues, as we may also construct queues from preferred operators~\cite{richter:helmert:icaps09,richter:westphal:jair10} or total orders of states~\cite{garrett:etal:ijcai16}.

\subsection{Portfolios}
Given that there is no single numeric planner configuration that performs best for all domains, the research community introduced portfolios and automatic configuration selection of planners, both of which combine standalone planners to maximise coverage over a diverse set of domains. 
Both portfolios and automatic configuration selection methods usually perform well on optimal and satisficing tracks of the International Planning Competitions.
Portfolios~\cite{helmert:etal:pal11} statically select a set of planners to run in sequence, each with a set amount of timeout such that the sum of their timeout is equal to the total timeout of the portfolio planner.
The partitioning of time resources for planners can be selected from training data or can simply be uniform~\cite{seipp:etal:icaps12}.
Similarly to portfolios, one may also just learn to choose a specific planner for a given domain~\cite{fawcett:etal:pal11,katz:etal:ipc18,ma:etal:aaai20}.
To reduce overfitting to benchmarks and to give a better understanding of the state of numeric planning domains, we opt to only use uniform portfolios in our experiments.

\section{Experiments}
\newcommand{\tablesize}{\small}
\renewcommand{\cellfontsize}{\footnotesize}
\renewcommand{\arraystretch}{1.1}
\renewcommand{\tabcolsep}{0.1cm}
\begin{table}
    \centering
    \small
    \begin{tabularx}{\columnwidth}{YYYYYYYYYYYY}
% \toprule
\multicolumn{7}{c}{Numeric Heuristics} & 
\multicolumn{4}{c}{Novelty Heuristics} \\
\cmidrule(l){1-7}
\cmidrule(l){8-11}
\header{$\gc$} & 
\header{$\hgdc$} & 
\header{$\aibr$} & 
\header{$\hadd$} & 
\header{$\hradd$} & 
\header{$\hmrp$} & 
\header{$\hmrphj$} & 
%
% \header{$\hadd$} & 
\header{$\hwadd$} & 
\header{$\hiwadd$} & 
\header{$\hqbadd$} & 
\header{$\hiqbadd$} \\
%
% \midrule
\cmidrule(l){1-7}
\cmidrule(l){8-11}
\normalcell{117} & 
\normalcell{200} & 
\normalcell{119} & 
\normalcell{183} & 
\normalcell{171} & 
\normalcell{176} & 
\second{217} & 
%
% \third{183} & 
\normalcell{178} & 
\normalcell{181} & 
\third{215} & 
\first{236} \\
% \third{183} & 
% \normalcell{178} & 
% \normalcell{181} & 
% \second{215} & 
% \first{236} \\
% \bottomrule
\end{tabularx}

    \caption{
        Total coverage of single queue GBFS with a selection of ENHSP heuristics and $\hadd$ novelty heuristics.
        The top three scores are indicated by the cell colour intensity.
    }
    \label{tab:num_and_nov}
\end{table}

\begin{table*}[ht]
    \centering
    \small
    \begin{tabularx}{\textwidth}{YYYYYYYYYYYYYY}
% \toprule
&
\multicolumn{6}{c}{S GBFS} 
% \multicolumn{6}{c}{single-queue} 
&
\multicolumn{3}{c}{M GBFS} 
% \multicolumn{3}{c}{multi-queue} 
&
\multicolumn{3}{c}{P GBFS}
&
\multicolumn{1}{r}{SMT}
\\
\cmidrule(l){2-7}
\cmidrule(l){8-10}
\cmidrule(l){11-13}
\cmidrule(l){14-14}
%
% \cmidrule(l){1-6}
% \cmidrule(l){7-9}
% \cmidrule(l){10-12}
% \cmidrule(l){13-13}
%
\header{num. tasks} &
\header{$\hgdc$} & \header{$\hadd$} & \header{$\hmrphj$} & \header{$\hiqbgdc$} & \header{$\hiqbadd$} & \header{$\hiqbmrphj$} & \header{$\mqe$} & \header{$\mqeiqb$} & \header{$\mqeeiqb$} & \header{$\pfh$} & \header{$\pfn$} & \header{$\pfhn$} & \header{$\patty$} 
\\
% \midrule
% \cmidrule{1-13}
\cmidrule(l){1-1}
\cmidrule(l){2-7}
\cmidrule(l){8-10}
\cmidrule(l){11-13}
\cmidrule(l){14-14}
400 & 
\normalcell{200} & \normalcell{183} & \normalcell{217} & \normalcell{185} & \normalcell{236} & \normalcell{222} & \normalcell{261} & \normalcell{244} & \normalcell{274} & \third{290} & \second{292} & \first{315} & \normalcell{262} 
\\
% \bottomrule
\end{tabularx}

    \caption{
        Total coverage of single queue ($\SQ$), multi-queue ($\MQ$), and portfolio ($\PF$) GBFS configurations, and the \patty{} solver.
        The top three scores are indicated by the cell colour intensity.
    }
    \label{tab:coverage}
\end{table*}

We evaluate the effectiveness of novelty heuristics and multi-queue search for numeric planning on the Numeric Track of the 2023 International Planning Competition (NT-IPC)~\cite{arxer:scala:23}. 
Its benchmark suite consists of 20 domains with 20 problems each. 
Action effects are limited to either linear or simple effects.
Heuristics we consider are goal count $(\hgc)$, \hgdcname{} $(\hgdc)$, the additive interval-based relaxation $(\aibr)$~\cite{scala:etal:ecai16}, the subgoaling additive heuristic with $(\hradd)$ and without $(\hadd)$ redundant constraints~\cite{scala:etal:ijcai16}, and the multi-repetition relaxed plan heuristic ($\hmrp$) with successors restricted to states generated by up-to-jumping actions $(\hmrphj)$.
Given a novelty feature $f$, a heuristic $h$, and a novelty heuristic $n$, we write $h_{\gen{f,n}}$ for the corresponding heuristic.
The first entry in $f(\seqState{T}, s)$ is given by the sequence of previously evaluated states during search.
We let $3h$ denote the list of top 3 performing ENHSP heuristics $\hgdc$, $\hadd$, and $\hmrphj$, $3n$ the list of their corresponding novelty heuristics $\hiqbgdc$, $\hiqbadd$, and $\hiqbmrphj$ with $k=2$, and $3h\concat 3n$ the concatenation of these two lists.
We choose the novelty heuristic extension $\gen{\B, \QB}$ as it is the best forming novelty heuristic overall.
Then $\MQ(\cdot)$ denotes a multi-queue search with a queue for each input heuristic, and $\PF(\cdot)$ a portfolio with a uniform partitioning of time for each input heuristic.
$\patty$ denotes the SMT planner by \citet{cardellini:etal:aaai24}.
All experiments are run with a 600 second timeout and 8GB memory limit on a single Intel Xeon 3.2 GHz CPU core.
To help with the evaluation of the effectiveness of the outlined search techniques, we perform experiments to answer the following questions.
{We refer to the appendix for more detailed plots and tables.}

\subsubsection*{How effective is $\hgdc$?}
From Tab.~\ref{tab:num_and_nov}, we notice that $\hgdc$ is the best performing heuristic behind $\hmrphj$.
This can be attributed to its fast evaluation speed rather than its informativeness as it generally expands more nodes than other heuristics on problems which both $\hgdc$ and another heuristic solves.
Exceptions to this rule are the fo-farmland and tpp domains in which delete relaxation heuristics perform poorly.
Nevertheless it generally expands far fewer nodes than $\hgc$.
Unfortunately, $\hgdc$ returns the worst plan length on almost all problems compared to all other heuristics.

\subsubsection*{What is the best novelty heuristic for numeric planning?}
From Tab.~\ref{tab:num_and_nov}, we notice that $\hiqbadd$ is the best performing novelty heuristic in terms of coverage.
Observing the performance of configurations, the effect of choosing $\QB$ over $\PN$ for novelty heuristic is greater than choosing $\B$ over $\A$ for the novelty feature.
This is supported by the fewer expansions of the $\QB$ variants compared to their $\PN$ counterparts over almost all domains.
On the other hand, the comparison of plan length depends on the domain.
The performance improvement over the base heuristic also depends on the domain.
We lastly note that choosing $k=2$ almost always provides better performance than $k=1$ for all configurations of novelty heuristics.

\subsubsection*{Is multi-queue search or using portfolios better for combining heuristics?}
Multi-queue search saves computation by reducing redundant node expansions and is able to make use of the exploration vs. exploitation paradigm when using multiple diverse heuristics.
On the other hand, portfolios make use of the fact that some heuristics perform better on some domains than others.
From Tab.~\ref{tab:coverage}, we notice that portfolios overall achieve higher coverage even when using the simple uniform partitioning scheduling.
The $\pfhn$ configuration achieves strictly better coverage than $\mqeeiqb$ on 9 domains, and strictly lower coverage on 5 domains.
This suggests that some domains can be solved efficiently by a single heuristic, while on other domains, we require the exploration effect of multi-queue search when none of the heuristics perform well.
Furthermore, it depends on the domain whether the best heuristic or multi-queue search expands fewer nodes and returns better quality plans.

\subsubsection*{Which domains are more suited to constraint-based solvers and which to heuristic search?}
We note that the best performing constraint-based solver \patty{} solves strictly more problems than $\pfhn$ on 5 domains, and strictly fewer problems on 9 domains.
The problems in which \patty{} performs well generally have a high ratio of numeric variables and short rolled up plans, such as the fo-counters and fo-sailing domains.
These problems have infinitely large state spaces and it is more likely for search-based methods to get lost.
However, search-based methods perform better on domains which require long sequential plans or traversal over maps with many locations, such as delivery, drone, ext-plant-watering, markettrader and tpp.
This is because SMT encodings may require many decision variables to represent problems with complex, sequential plans.
However, we note that \patty{} generally returns poorer quality plans with plan lengths up to an order of magnitude longer than search planners.

\section{Conclusion}
We have ported search techniques from classical planning to boost the performance of satisficing numeric planning solvers.
We have formalised novelty heuristics for numeric planning, constructed a simple but powerful \hgdcname{} heuristic, and extended the ENHSP planner with multi-queue search and naive portfolios.
Experiments demonstrate the effectiveness of these techniques on the recent Numeric Track of the 2023 International Planning Competition (NT-IPC).
Results also suggest a need for more diverse numeric planning benchmarks, as 78.8\% of the problems from the NT-IPC are solved by our new $\pfhn$ configuration in the ENHSP planner within 5 minutes, and 89.2\% of the problems by either $\pfhn$ or \patty{}.

\clearpage
\small
\section*{Acknowledgments}
The authors thank the reviewers for the helpful comments and suggestions, most notably with the naming of the Manhattan distance heuristic by connecting the computation of errors to the $L_1$ metric.
We also thank Enrico Scala for providing the source code for ENHSP.
The computing resources for the project was supported by the Australian Government through the National Computational Infrastructure (NCI) under the ANU Startup Scheme
This work was supported by Australian Research Council grant DP220103815, by the Artificial and Natural Intelligence Toulouse Institute (ANITI) under the grant agreement ANR-19-PI3A-000, and by the European Union's Horizon Europe Research and Innovation program under the grant agreement TUPLES No. 101070149.

\bibliography{references.bib}
\normalsize

\clearpage
\appendix
\onecolumn

\section{Benchmark Domains}
\newcolumntype{Y}{>{\raggedleft\arraybackslash}X}
\begin{table}[h]
    \centering
    \begin{tabularx}{\columnwidth}{X X Y Y Y}
        \toprule
        Domain & Effects & $|A|$ & $|B|$ & $|N|$ \\
        \midrule
        block-grouping & simple & 4 & 0 & 6 \\
        counters & simple & 2 & 0 & 2 \\
        delivery & simple & 5 & 7 & 4 \\
        drone & simple & 8 & 1 & 14 \\
        expedition & simple & 4 & 2 & 3 \\
        ext-plant-watering & simple & 10 & 0 & 12 \\
        farmland & simple & 2 & 1 & 2 \\
        fo-counters & linear & 4 & 0 & 4 \\
        fo-farmland & linear & 3 & 2 & 3 \\
        fo-sailing & linear & 10 & 2 & 4 \\
        hydropower & simple & 3 & 7 & 4 \\
        markettrader & simple & 4 & 2 & 7 \\
        mprime & simple & 4 & 3 & 2 \\
        pathwaysmetric & simple & 6 & 6 & 14 \\
        rover & simple & 10 & 26 & 2 \\
        sailing & simple & 8 & 1 & 3 \\
        settlersnumeric & simple & 25 & 19 & 7 \\
        sugar & simple & 13 & 11 & 20 \\
        tpp & linear & 3 & 1 & 6 \\
        zenotravel & simple & 5 & 2 & 8 \\
        \bottomrule
    \end{tabularx}
    \caption{NT-IPC~\cite{arxer:scala:23} domains used in the experiments. $|A|$, $|B|$, and $|N|$ are the number of action schemata, predicates and functions in each domain, respectively.}
\end{table}

\section{Scatter Plots}
Please refer to Figures~\ref{fig:best-h-vs-mqeeiqb}, \ref{fig:hgdc-vs-heuristics}, and \ref{fig:hiqbadd-vs-heuristics} for scatter plots comparing various planners in terms of plan length and node expansions, and Fig.~\ref{fig:legend} for the legend of domains.

\newcommand{\figscale}{0.32}
\newcommand{\legscale}{0.6}

\begin{figure}[ht!]
    \centering
    \begin{minipage}{.33\textwidth}
      \centering
      \raisebox{-0.43\height}{\fbox{
        \includegraphics[height=\legscale\textwidth]{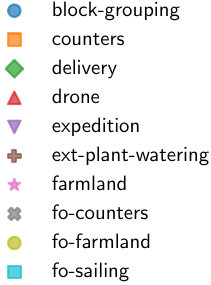}
        \includegraphics[height=\legscale\textwidth]{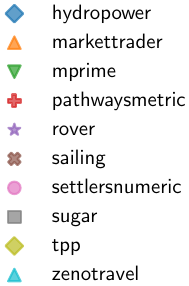}
      }}
      \captionof{figure}{Legend of domains.}
      \label{fig:legend}
    \end{minipage}%
    \hfill
    \begin{minipage}{.65\textwidth}
      \centering
      \includegraphics[width=\figscale\textwidth]{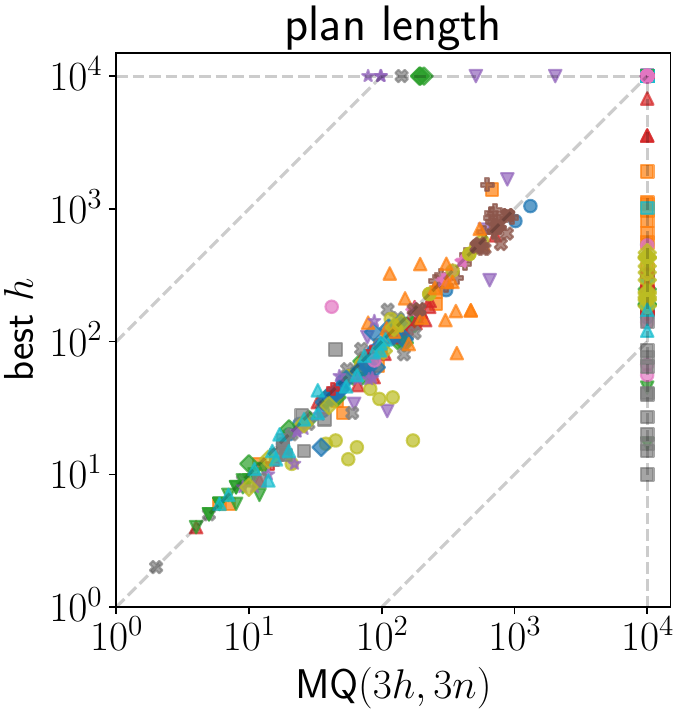}
      {\includegraphics[width=\figscale\textwidth]{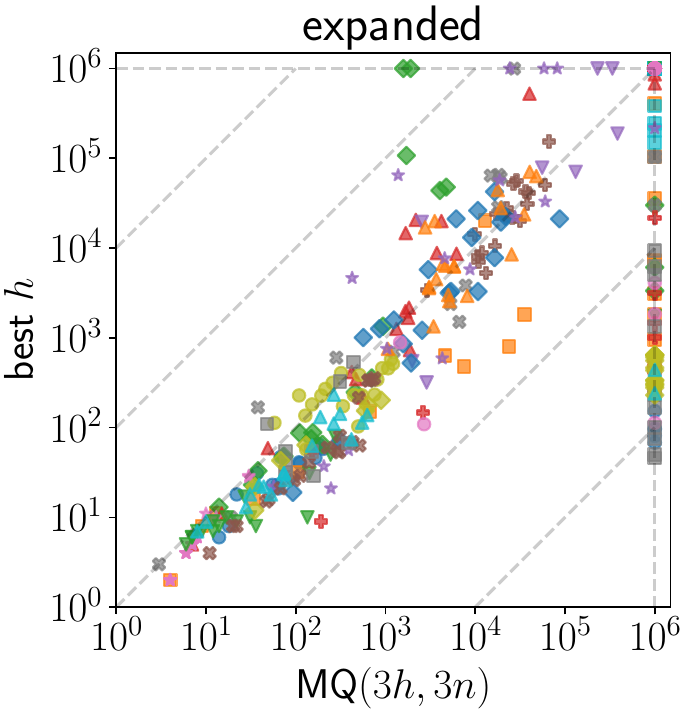}}
      {\includegraphics[width=\figscale\textwidth]{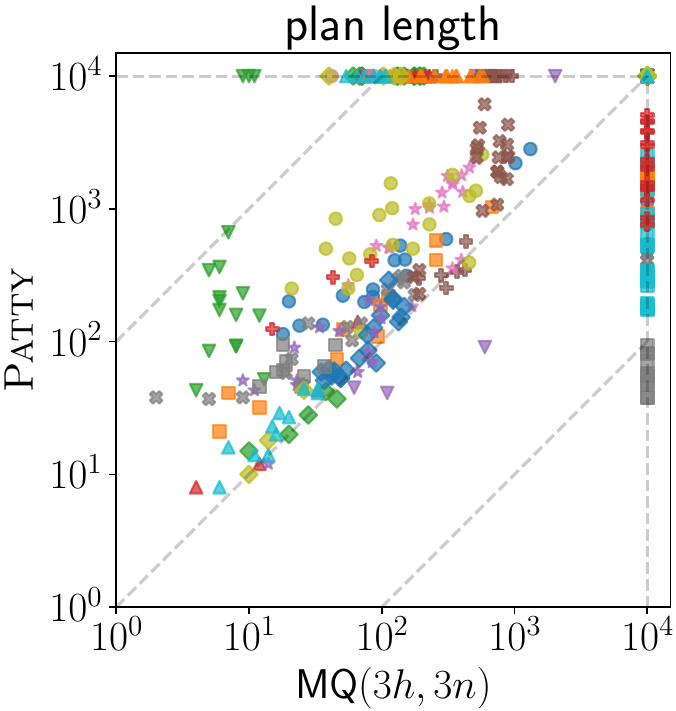}}
      \captionof{figure}{
        Left and center: $\text{best }h$, the best SQ GBFS planner for each problem, vs $\mqeeiqb$ on sequential plan length and number of expanded nodes. 
        Right: $\patty$ vs $\mqeeiqb$ on sequential plan length. 
        Points on the top left triangle favour $\mqeeiqb$ and points on the bottom right favour $\text{best }h$ and $\patty$ on the respective plots.
      }
      \label{fig:best-h-vs-mqeeiqb}
    \end{minipage}
    \end{figure}

\renewcommand{\figscale}{0.196}
\begin{figure*}[h!]
    \centering
    \raisebox{-0.5\height}{\includegraphics[width=\figscale\textwidth]{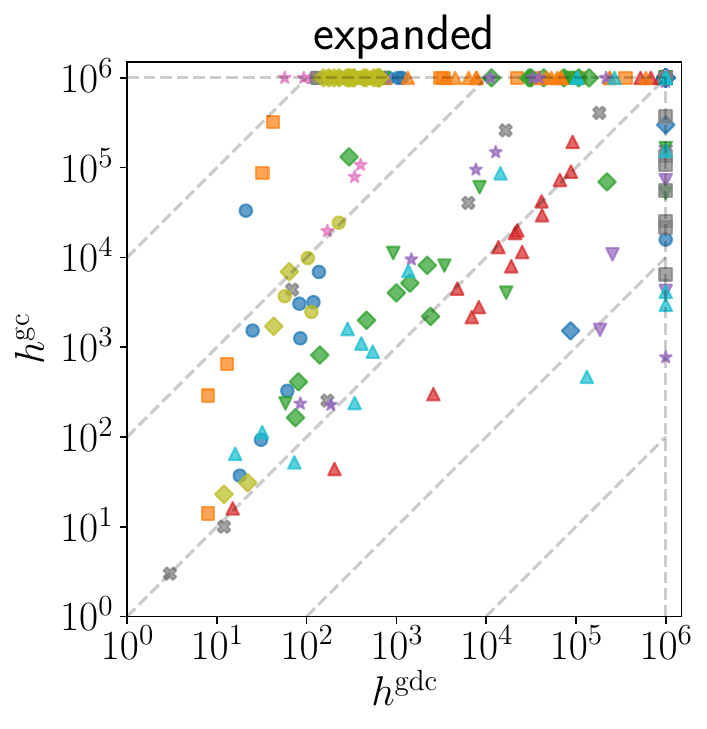}}
    \hfill
    \raisebox{-0.5\height}{\includegraphics[width=\figscale\textwidth]{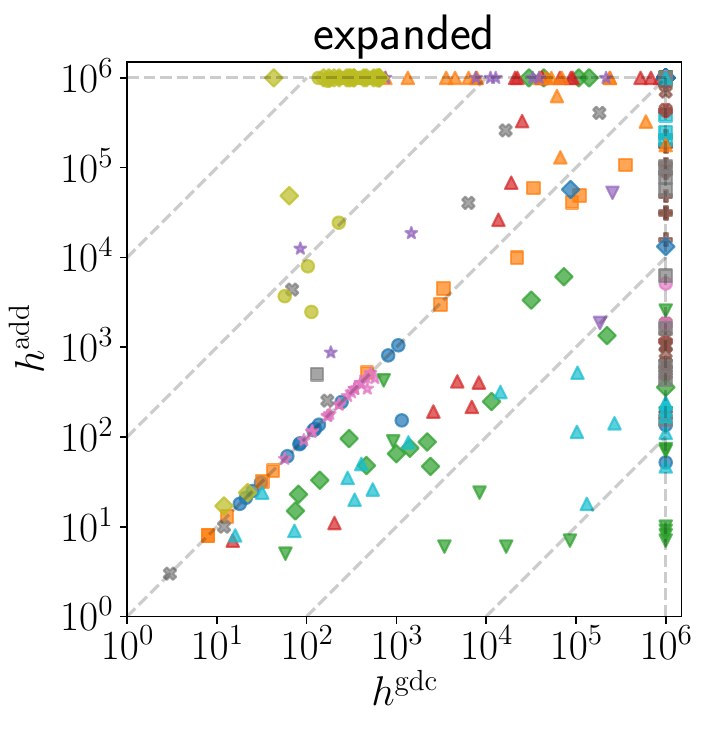}}
    \hfill
    \raisebox{-0.5\height}{\includegraphics[width=\figscale\textwidth]{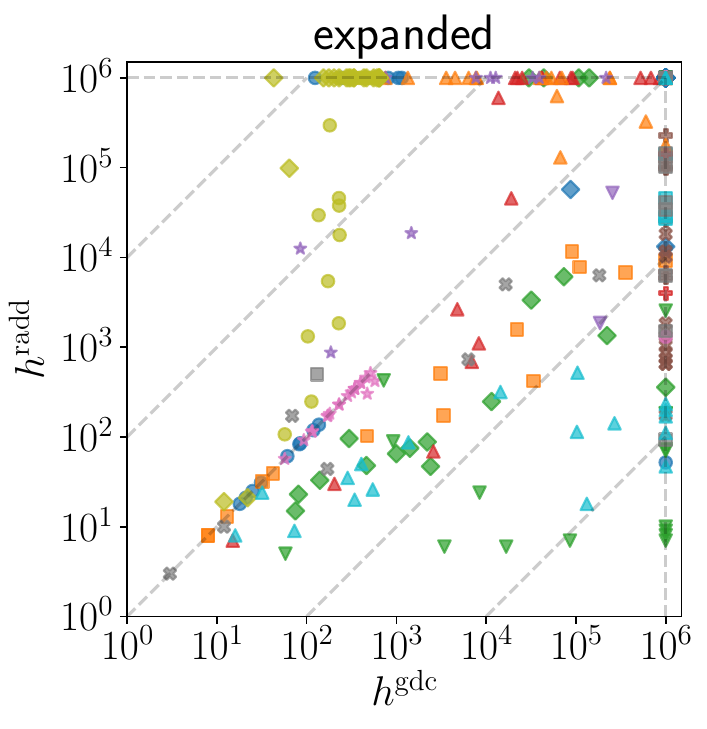}}
    \hfill
    \raisebox{-0.5\height}{\includegraphics[width=\figscale\textwidth]{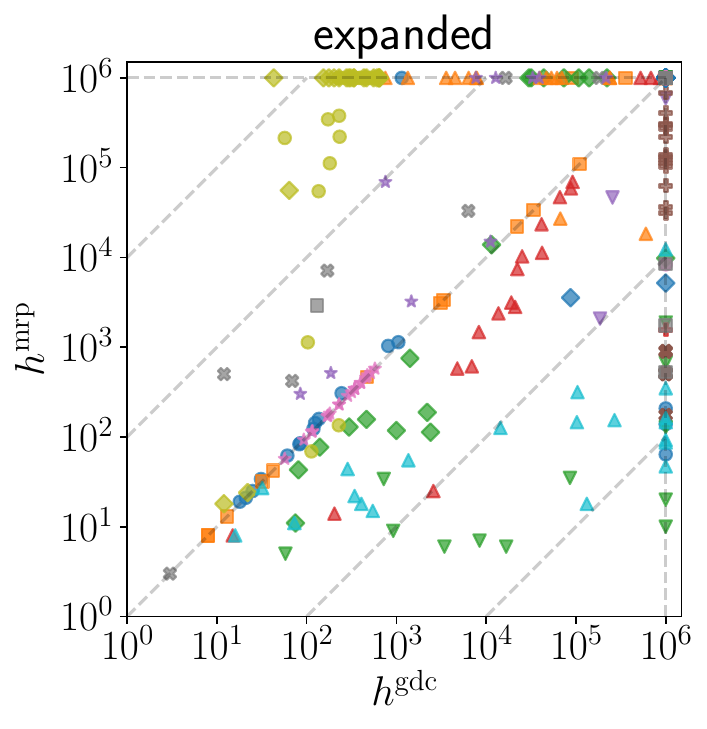}}
    \hfill
    \raisebox{-0.5\height}{\includegraphics[width=\figscale\textwidth]{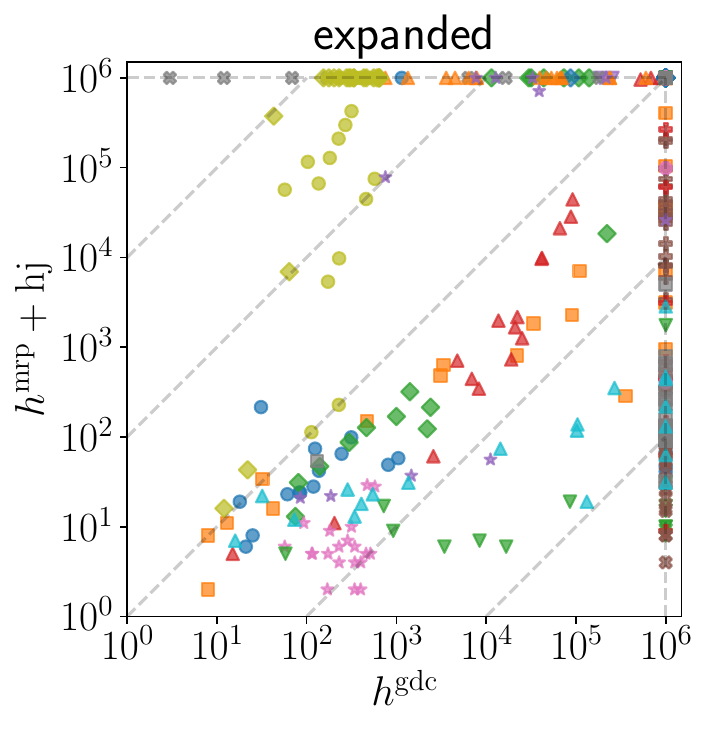}}
    \hfill
    \raisebox{-0.5\height}{\includegraphics[width=\figscale\textwidth]{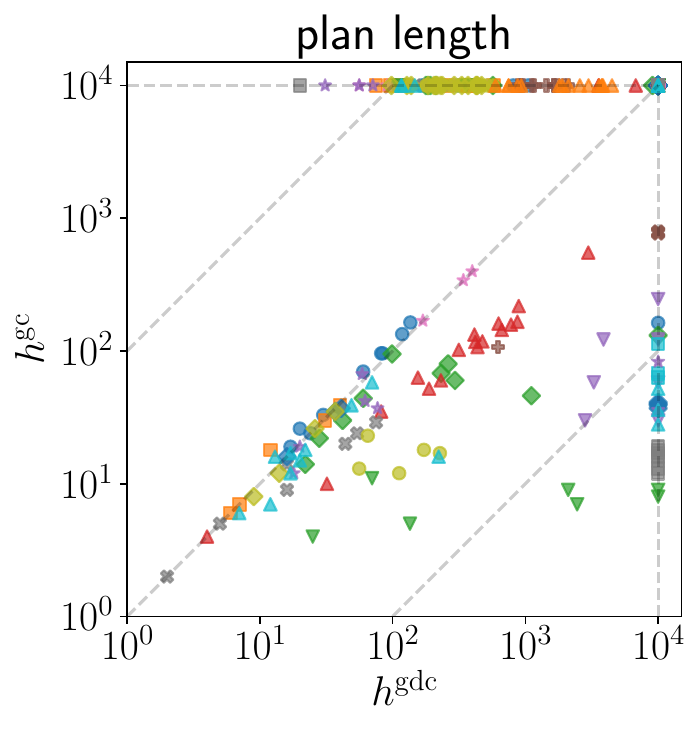}}
    \hfill
    \raisebox{-0.5\height}{\includegraphics[width=\figscale\textwidth]{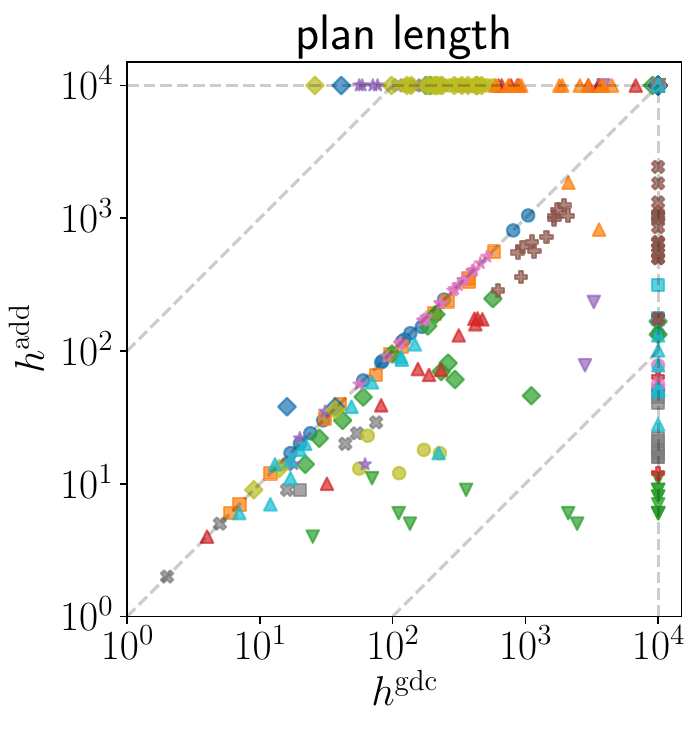}}
    \hfill
    \raisebox{-0.5\height}{\includegraphics[width=\figscale\textwidth]{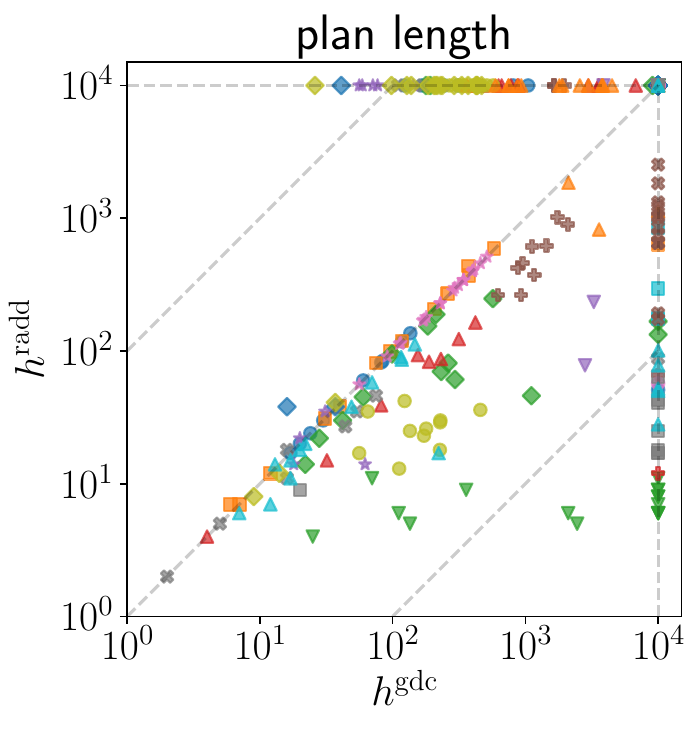}}
    \hfill
    \raisebox{-0.5\height}{\includegraphics[width=\figscale\textwidth]{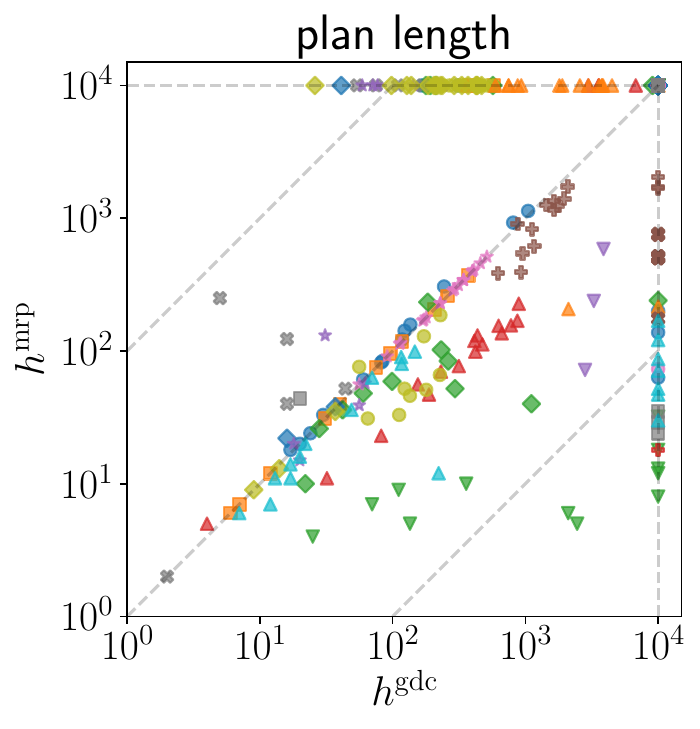}}
    \hfill
    \raisebox{-0.5\height}{\includegraphics[width=\figscale\textwidth]{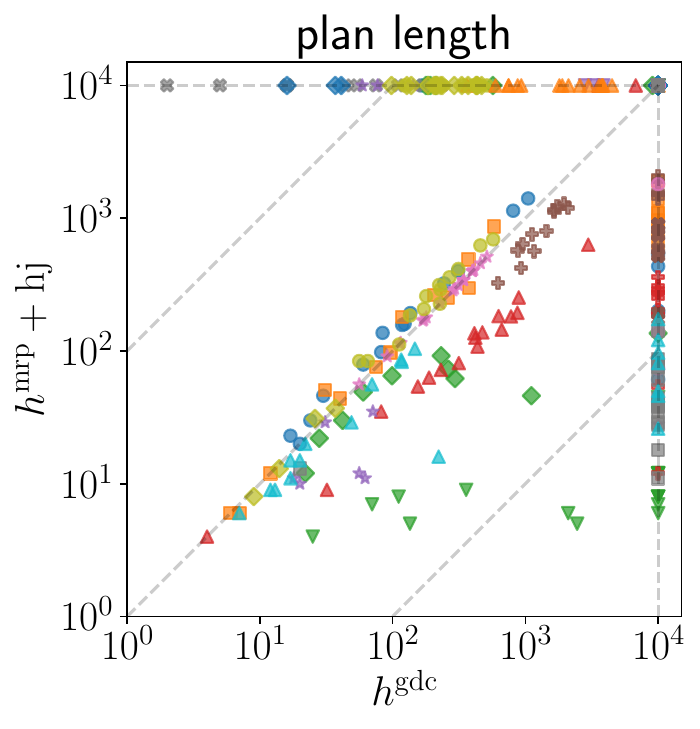}}
    \caption{
        $\hgdc$ ($x$-axis) vs. various heuristics ($y$-axis) in terms of plan length and number of nodes expanded during search.
        Top left points benefit $\hgdc$ and bottom right points the $y$-axis configuration.
    }
    \label{fig:hgdc-vs-heuristics}
\end{figure*}

\renewcommand{\figscale}{0.2}
\begin{figure*}[h!]
    \centering
    \raisebox{-0.5\height}{\includegraphics[width=\figscale\textwidth]{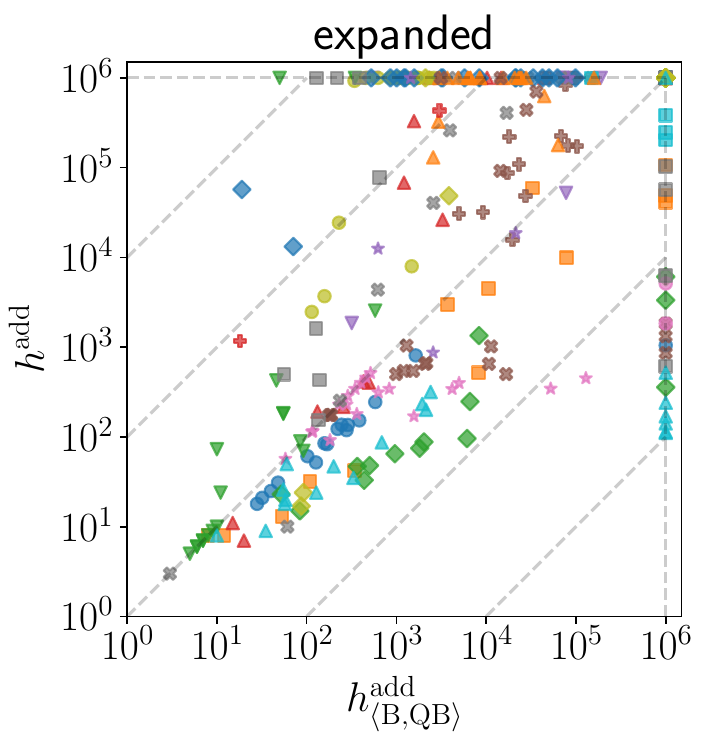}}
    \hfill
    \raisebox{-0.5\height}{\includegraphics[width=\figscale\textwidth]{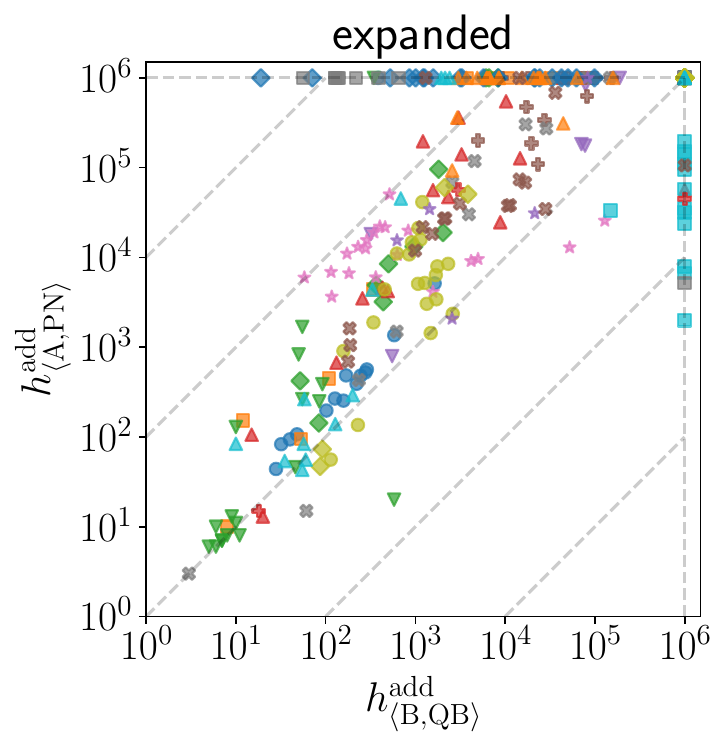}}
    \hfill
    \raisebox{-0.5\height}{\includegraphics[width=\figscale\textwidth]{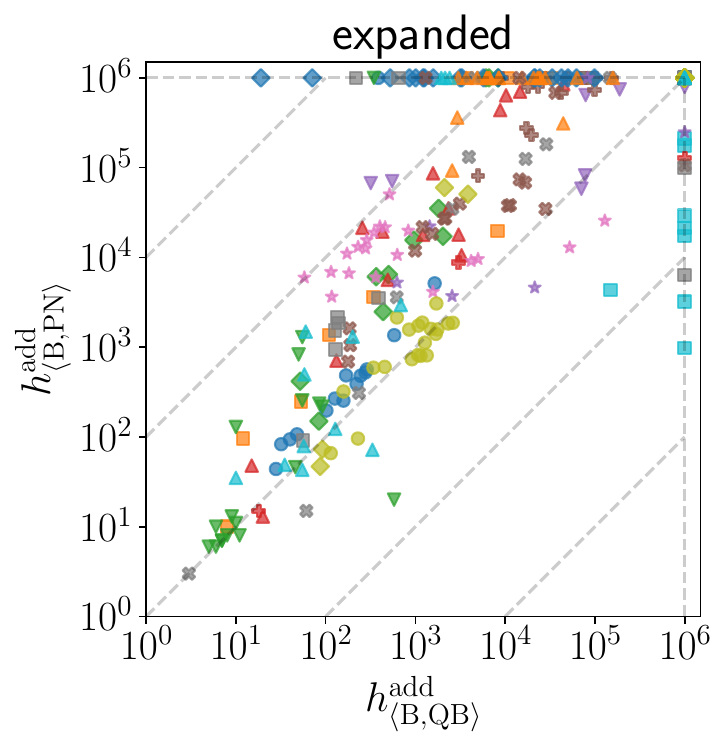}}
    \hfill
    \raisebox{-0.5\height}{\includegraphics[width=\figscale\textwidth]{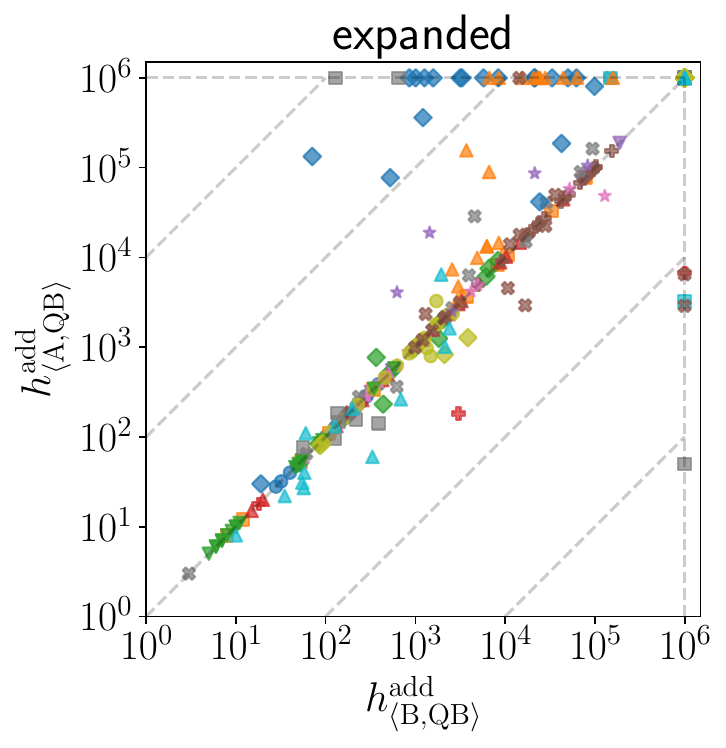}}
    \hfill
    \raisebox{-0.5\height}{\includegraphics[width=\figscale\textwidth]{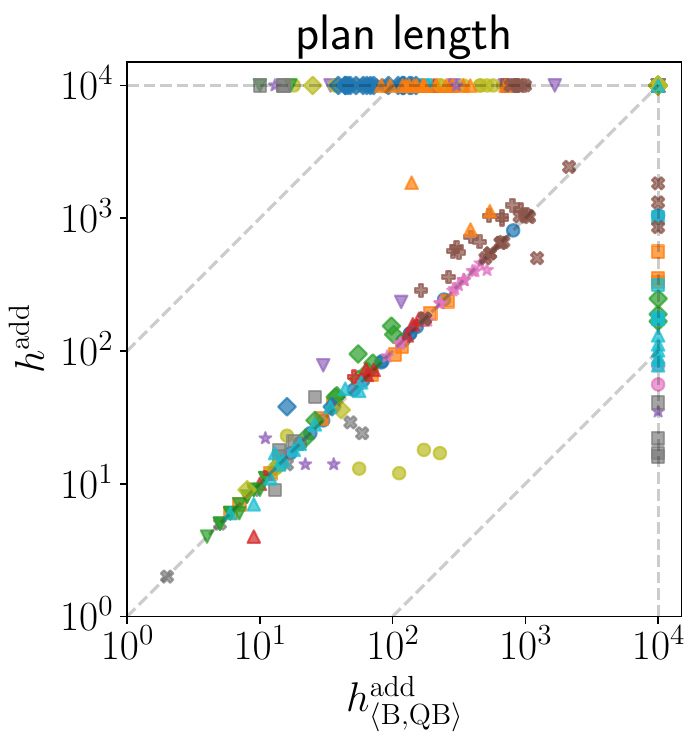}}
    \hfill
    \raisebox{-0.5\height}{\includegraphics[width=\figscale\textwidth]{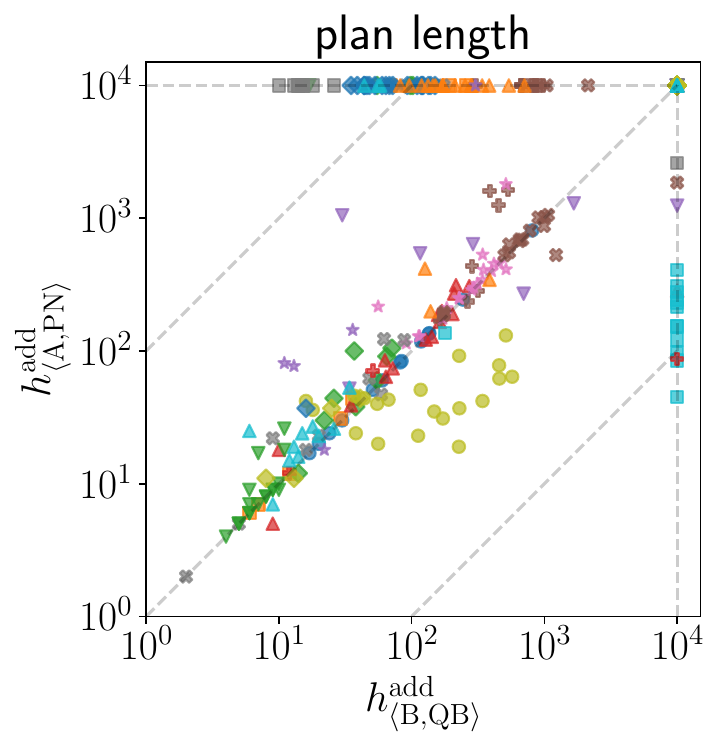}}
    \hfill
    \raisebox{-0.5\height}{\includegraphics[width=\figscale\textwidth]{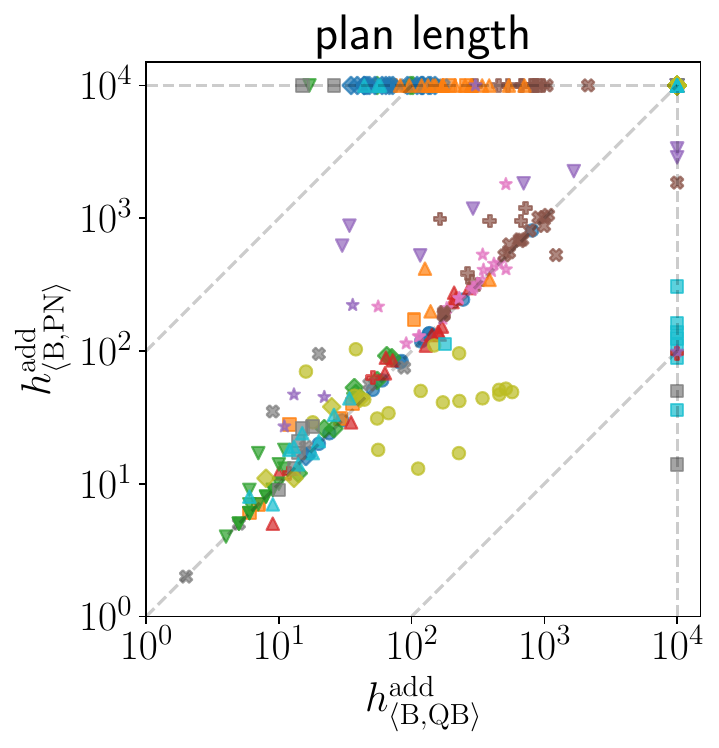}}
    \hfill
    \raisebox{-0.5\height}{\includegraphics[width=\figscale\textwidth]{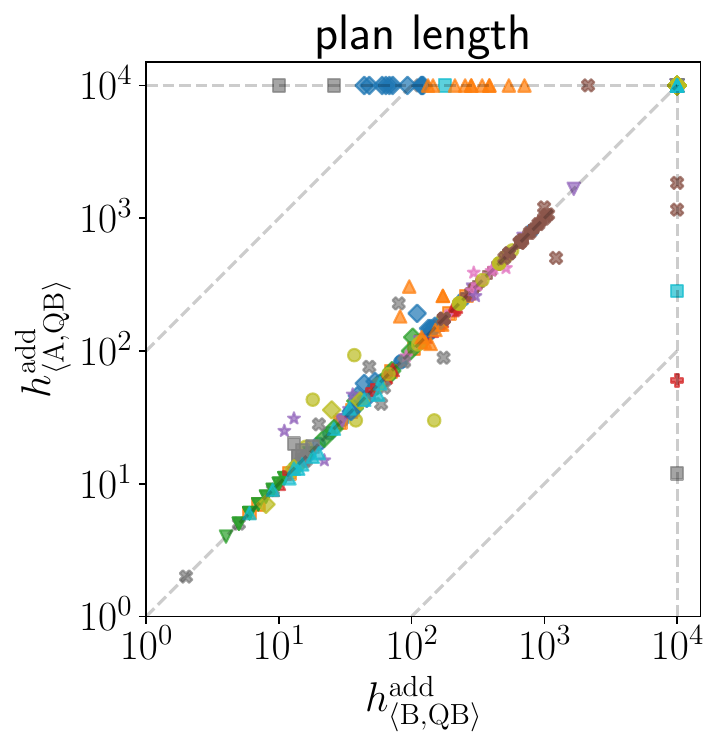}}
    \caption{
        $\hiqbadd$ ($x$-axis) vs. $\hadd$ and other novelty extensions ($y$-axis) in terms of plan length and number of nodes expanded during search.
        Top left points benefit $\hiqbadd$ and bottom right points the $y$-axis configuration.
    }
    \label{fig:hiqbadd-vs-heuristics}
\end{figure*}

\clearpage
\section{Full coverage details}
The total coverage of planners over each domain is shown in Table~\ref{tab:coverage_all}.
Almost all domains have all problems solved by at least one planner (block-grouping, counters, drone, ext-plant-watering, farmland, fo-counters, fo-farmland, fo-sailing, hydropower, markettrader, pathwaysmetric, sailing, sugar, tpp, and zenotravel).
This can either be attributed to small problem sizes, a planner being designed or motivated primarily for the domain, or the problem becomes trivial when not considering plan quality, such as farmland, markettrader or tpp.

The two domains which are difficult for the selected planners are expedition and settlersnumeric.
Expedition is a resource transportation problem with the problem that the agent(s) must go back and forth along the transportation path due to having limited fuel and carrying capacity. The considered heuristics cannot reason about the requirement of going back and forth due to being derived from delete relaxation.
Settlersnumeric is modelling a civilisation-like game which requires managing resources to build town structures. There is a large number of predicates in this problem and again delete relaxation heuristics cannot detect long range dependencies of resources, e.g. wood is needed to build a saw mill which in turn also needs wood to create timber. The domain is solved well by LP-RPG~\cite{coles:etal:jair13} whose paper introduced and was motivated by the domain.

\renewcommand{\cellfontsize}{\small}
\begin{table*}[h]%[ht]
    \centering
    \begin{tabularx}{\textwidth}{lYYYYYYYYYYYYYY}
\toprule
&
\multicolumn{6}{c}{SQ GBFS} 
&
\multicolumn{3}{c}{MQ GBFS} 
&
\multicolumn{3}{c}{PF GBFS}
&
\multicolumn{1}{r}{SMT}
\\
\cmidrule(l){2-7}
\cmidrule(l){8-10}
\cmidrule(l){11-13}
\cmidrule(l){14-14}
domain & \header{$\hgdc$} & \header{$\hadd$} & \header{$\hmrphj$} & \header{$\hiqbgdc$} & \header{$\hiqbadd$} & \header{$\hiqbmrphj$} & \header{$\mqe$} & \header{$\mqeiqb$} & \header{$\mqeeiqb$} & \header{$\pfh$} & \header{$\pfn$} & \header{$\pfhn$} & \header{$\patty$} & \header{best}\\
% \midrule
\cmidrule(lr){1-1}
\cmidrule{2-14}
\cmidrule(l){15-15}
\normalcell{block-grouping }  & \normalcell{15} & \normalcell{16} & \second{19} & \normalcell{15} & \normalcell{15} & \second{19} & \normalcell{16} & \normalcell{12} & \normalcell{15} & \normalcell{18} & \normalcell{16} & \normalcell{17} & \first{20} & \normalcell{20} \\
\normalcell{counters }  & \normalcell{13} & \normalcell{13} & \first{20} & \normalcell{10} & \normalcell{10} & \first{20} & \normalcell{14} & \normalcell{9} & \normalcell{11} & \first{20} & \normalcell{19} & \first{20} & \first{20} & \normalcell{20} \\
\normalcell{delivery }  & \third{17} & \normalcell{14} & \normalcell{10} & \normalcell{14} & \normalcell{11} & \normalcell{10} & \first{19} & \normalcell{10} & \third{17} & \second{18} & \normalcell{12} & \normalcell{16} & \normalcell{5} & \normalcell{19} \\
\normalcell{drone }  & \first{20} & \normalcell{11} & \normalcell{17} & \normalcell{5} & \normalcell{15} & \normalcell{17} & \normalcell{16} & \normalcell{18} & \normalcell{17} & \second{19} & \normalcell{16} & \second{19} & \normalcell{4} & \normalcell{20} \\
\normalcell{expedition }  & \normalcell{3} & \normalcell{2} & \zerocell{0} & \normalcell{3} & \second{6} & \zerocell{0} & \normalcell{3} & \normalcell{5} & \first{8} & \normalcell{3} & \second{6} & \second{6} & \normalcell{3} & \normalcell{8} \\
\normalcell{ext-plant-watering }  & \normalcell{12} & \normalcell{12} & \first{20} & \normalcell{1} & \normalcell{19} & \first{20} & \normalcell{17} & \first{20} & \first{20} & \first{20} & \first{20} & \first{20} & \normalcell{6} & \normalcell{20} \\
\normalcell{farmland }  & \first{20} & \first{20} & \first{20} & \first{20} & \first{20} & \first{20} & \first{20} & \first{20} & \first{20} & \first{20} & \first{20} & \first{20} & \first{20} & \normalcell{20} \\
\normalcell{fo-counters }  & \normalcell{9} & \normalcell{7} & \zerocell{0} & \second{15} & \normalcell{12} & \zerocell{0} & \normalcell{9} & \third{14} & \third{14} & \normalcell{8} & \third{14} & \normalcell{10} & \first{20} & \normalcell{20} \\
\normalcell{fo-farmland }  & \first{20} & \normalcell{5} & \normalcell{13} & \first{20} & \first{20} & \normalcell{14} & \first{20} & \first{20} & \first{20} & \first{20} & \first{20} & \first{20} & \first{20} & \normalcell{20} \\
\normalcell{fo-sailing }  & \zerocell{0} & \second{4} & \zerocell{0} & \zerocell{0} & \normalcell{1} & \zerocell{0} & \normalcell{3} & \zerocell{0} & \zerocell{0} & \normalcell{3} & \normalcell{1} & \second{4} & \first{20} & \normalcell{20} \\
\normalcell{hydropower }  & \normalcell{3} & \normalcell{2} & \zerocell{0} & \normalcell{18} & \first{20} & \zerocell{0} & \normalcell{3} & \first{20} & \first{20} & \normalcell{2} & \first{20} & \first{20} & \first{20} & \normalcell{20} \\
\normalcell{markettrader }  & \normalcell{17} & \normalcell{4} & \zerocell{0} & \first{20} & \first{20} & \zerocell{0} & \normalcell{15} & \first{20} & \first{20} & \normalcell{17} & \first{20} & \normalcell{19} & \zerocell{0} & \normalcell{20} \\
\normalcell{mprime }  & \normalcell{7} & \normalcell{16} & \normalcell{16} & \normalcell{12} & \second{18} & \normalcell{16} & \first{19} & \second{18} & \normalcell{17} & \normalcell{17} & \second{18} & \second{18} & \normalcell{14} & \normalcell{19} \\
\normalcell{pathwaysmetric }  & \zerocell{0} & \normalcell{2} & \normalcell{12} & \zerocell{0} & \normalcell{2} & \second{13} & \normalcell{3} & \normalcell{2} & \normalcell{3} & \normalcell{12} & \second{13} & \normalcell{12} & \first{20} & \normalcell{20} \\
\normalcell{rover }  & \normalcell{10} & \normalcell{4} & \normalcell{7} & \normalcell{12} & \normalcell{5} & \normalcell{7} & \third{13} & \normalcell{12} & \first{16} & \normalcell{11} & \normalcell{12} & \normalcell{12} & \first{16} & \normalcell{16} \\
\normalcell{sailing }  & \zerocell{0} & \normalcell{18} & \first{20} & \zerocell{0} & \normalcell{17} & \first{20} & \first{20} & \normalcell{19} & \first{20} & \first{20} & \first{20} & \first{20} & \first{20} & \normalcell{20} \\
\normalcell{settlersnumeric }  & \zerocell{0} & \normalcell{2} & \normalcell{3} & \zerocell{0} & \zerocell{0} & \second{4} & \second{4} & \zerocell{0} & \normalcell{2} & \second{4} & \normalcell{3} & \first{5} & \zerocell{0} & \normalcell{5} \\
\normalcell{sugar }  & \normalcell{1} & \normalcell{9} & \normalcell{16} & \zerocell{0} & \normalcell{8} & \third{18} & \normalcell{12} & \normalcell{9} & \normalcell{9} & \third{18} & \third{18} & \second{19} & \first{20} & \normalcell{20} \\
\normalcell{tpp }  & \first{20} & \normalcell{3} & \normalcell{4} & \normalcell{5} & \normalcell{4} & \normalcell{4} & \normalcell{17} & \normalcell{4} & \normalcell{7} & \first{20} & \normalcell{4} & \first{20} & \normalcell{3} & \normalcell{20} \\
\normalcell{zenotravel }  & \normalcell{13} & \normalcell{19} & \first{20} & \normalcell{15} & \normalcell{13} & \first{20} & \normalcell{18} & \normalcell{12} & \normalcell{18} & \first{20} & \first{20} & \normalcell{18} & \normalcell{11} & \normalcell{20} \\
% \midrule
\cmidrule(lr){1-1}
\cmidrule{2-14}
\cmidrule(l){15-15}
\normalcell{sum}  & \normalcell{200} & \normalcell{183} & \normalcell{217} & \normalcell{185} & \normalcell{236} & \normalcell{222} & \normalcell{261} & \normalcell{244} & \normalcell{274} & \third{290} & \second{292} & \first{315} & \normalcell{262} & \normalcell{367} \\
\bottomrule
\end{tabularx}

    \caption{
        Coverage of various configurations of single queue ($\SQ$), multi-queue ($\MQ$), and portfolio ($\PF$) GBFS, and the \patty{} solver on the NT-IPC domains.
        The rightmost column records the highest coverage for each domain.
        The top scores for each row except the rightmost column are indicated by the cell colour intensity with the top score being highlighted in bold.
    }
    \label{tab:coverage_all}
\end{table*}

\end{document}